%% file: main.tex
\documentclass[
twocolumn,
]{ceurart}

\usepackage{listings}
\usepackage{hyperref}
\hypersetup{
    colorlinks=true,
    linkcolor=blue,
    filecolor=magenta,      
    urlcolor=blue,
    }

\usepackage{tikz}
\newcommand\encircle[1]{%
  \tikz[baseline=(X.base)] 
    \node (X) [draw, shape=circle, inner sep=0] {\strut #1};}

\usepackage{chngcntr}
\usepackage{todonotes}
\begin{document}

\copyrightyear{2022}
\copyrightclause{Copyright for this paper by its authors.
  Use permitted under Creative Commons License Attribution 4.0
  International (CC BY 4.0).}

\conference{AIMLAI @ CIKM'22: Advances in Interpretable Machine Learning and Artificial Intelligence,
  October 21, 2022, Atlanta, GA}

\title{Explainer Divergence Scores (EDS): Some Post-Hoc Explanations May be Effective for Detecting Unknown Spurious Correlations}

\author[1,2]{Shea Cardozo}[%
email=shea.cardozo@tenyks.ai,
]
\fnmark[1]
\address[1]{Tenyks}
\address[2]{University of Toronto}

\author[1, 2]{Gabriel Islas Montero}[%
email=gabriel.montero@tenyks.ai,
]
\fnmark[1]

\author[1, 3]{Dmitry Kazhdan}[%
email=dmitry.kazhdan@tenyks.ai,
]
\address[3]{University of Cambridge}

\author[1]{Botty Dimanov}[%
email=botty.dimanov@tenyks.ai,
]

\author[1]{Maleakhi Wijaya}[%
email=maleakhi.wijaya@tenyks.ai,
]

\author[3]{Mateja Jamnik}[%
email=mateja.jamnik@cl.cam.ac.uk,
]

\author[3]{Pietro Lio}[%
email=pl219@cam.ac.uk,
]

\fntext[1]{Equal Contribution}

\begin{abstract}
    Recent work has suggested post-hoc explainers might be ineffective for detecting spurious 
    correlations in Deep Neural Networks (DNNs). However, we show there are serious weaknesses with the existing evaluation frameworks for this setting. Previously proposed metrics are extremely difficult to interpret and are not directly comparable between explainer methods. To alleviate these constraints, we propose a new evaluation methodology, Explainer Divergence Scores (EDS),  grounded in an information theory approach to evaluate explainers.
    
    EDS is easy to interpret and naturally comparable across explainers. We use our methodology to compare the  detection performance of three different explainers - 
    feature attribution methods, influential examples and concept extraction, on two different image datasets. We discover post-hoc explainers often contain substantial information about a DNN's dependence on spurious artifacts, but in ways often imperceptible to human users. This suggests the need for new techniques that can use this information to better detect a DNN's reliance on spurious correlations.
    \end{abstract}
    

    \begin{keywords}
    explainability \sep
    interpretability \sep
    XAI \sep
    spurious correlations \sep
    explainer evaluation \sep
    post-hoc explanations \sep
    shortcut learning
    \end{keywords}

\maketitle

\section{Introduction}

Spurious correlations pose a serious risk to the application of Deep Neural Networks (DNNs), especially in critical applications, such as medical imaging and security~\cite{chouldechova2016fairpredict, doshi2017towards, datta2014automated, buolamwini2018gender}.
%\todo{cite Clever Hans papers, or use these examples from [13, 39, 10, 94, 32, 20, 17, 79 https://arxiv.org/pdf/2204.02937.pdf}. 
This phenomenon, also known as shortcut learning or the Clever Hans Effect, is the result of DNN's tendency to overfit to subtle patterns that are difficult for a human user to identify. This causes trained models to form decision rules that fail to generalise~\cite{lapuschkin2019cleverhans, geirhos2020shortcut, sagawa2020overparam, mahmood2021guidedradiology}. 
% The phenomenon of spurious correlations has haunted the field of statistical learning since its inception. With the increasing presence of highly representative machine learning systems in many critical applications, identifying and fixing these spurious correlations has become a topic of increasing interest. Deep Neural Networks (DNNs) are known to be  prone to overfitting to spurious patterns, and are notoriously opaque to interpret and explain

Consequently, detecting a model's dependency on a spurious signal (or `model spuriousness') in computer vision tasks has become an active area of research. Explainable AI (XAI) methods have been proposed as a potential avenue to address this challenge~\cite{lapuschkin2019cleverhans, geirhos2020shortcut,  adebayo2020debugging, han2020explaining}
. One of these methods, post-hoc explanations, aims to describe the inference process of a pre-trained DNN in a human-interpretable manner~\cite{doshi2017towards,dimanov2021interpretable}. 
%  An increasing number of techniques have been proposed in the emerging field of Explainable AI (XAI) to address this problem. These DNN 'explainers' aim to capture information about the inference process of the model useful for identifying model defects such as a reliance on spurious correlations. Despite the increasing usage and success of many of these explainers, there has yet to be an accepted systematic way to evaluate and compare them on a variety of different datasets, tasks and spurious signals. 

Past work has suggested human users may struggle to use post-hoc explanations to detect spurious signals if said spurious signal is not known ahead of time~\cite{adebayo2022posthoc}.

In this work, we ask deeper questions: 
\textit{Do post-hoc explanations contain any information that can be used to detect spurious signals even if the signal is \textbf{not known} ahead of time? If so, can we quantify and compare the amount of information different post-hoc explainers can extract?}

%While methods to evaluate the effectiveness of post-hoc explanations have been proposed in past work

% \todo{we are missing a few sentences here to link up the above with contributions, we can add an expanded version of the abstract }
In particular, we make the following contributions:
\begin{itemize}
    \item We propose Explainer Divergence Scores (EDS): a novel way to evaluate a post-hoc explainer's ability to detect spurious correlations based on an information theory foundation. 
    \item We show our method's effectiveness by evaluating and comparing three different types of post-hoc explainers - feature attribution methods \cite{sundararajan2017ig, ribeiro2016lime}, influential examples \cite{Garima2020estimating}, and concept extraction \cite{kazhdan2020cme} -  across multiple datasets \cite{dsprites17, 3dshapes18} and spurious artifacts. 
    \item  We compare the amount of information regarding the presence of a spurious signal \textbf{between} different post-hoc explainers, which existing approaches fail to address, and discover that post-hoc explainers contain a significant amount of information on model spuriousness. Since this information is frequently not visible to human users, our findings suggest that future research into post-hoc explanations should focus on discovering and utilising this information.
\end{itemize}

\input{Images/abstract_visual}

\section{Related Work}

\paragraph{Spurious Correlations} 

Spurious Correlations in DNNs have been the subject of a increasingly diverse body of work, with contributors analysing them through the lenses of distribution shift \cite{zhou2021distribution, sagawa2019distrobust}, shortcut learning \cite{geirhos2020shortcut} and causal inference \cite{arjovsky2019causal, moneda2021spurious-correlation-blogpost}. Spurious correlations have raised issues in areas as diverse as privacy \cite{leino2019membership}, fairness \cite{mehrabi2019fairness}, and adversarial attacks \cite{chen2017poisoning}. Recent work has focused on identifying where spurious correlations manifest and their properties, finding they often appear in practical settings~\cite{lapuschkin2019cleverhans, geirhos2020shortcut,xiao2020backgrounds, sagawa2020overparam, mahmood2021guidedradiology}.% For example, \citet{xiao2020backgrounds} find object detection models regularly depend on background information during inference, leading to spurious behaviour if the distribution of backgrounds change between training and inference. \citet{sagawa2020overparam} shows how models trained on the Waterbirds \cite{sagawa2019distrobust} and CelebA \cite{liu2015faceattributes} display spurious behaviour. \citet{yang2022rare} find DNN spuriousness can be induced with only a handful of spurious training examples. 

\paragraph{Post-Hoc Explainers}

Post-hoc explainability methods generate explanations of the inference process of an arbitrary trained DNN. Numerous post-hoc explainers have been proposed.

Feature attribution methods or `heatmaps' in Computer Vision domains, measure the effect of each individual input (e.g., pixel) on the output of a DNN by either leveraging input perturbation \cite{ribeiro2016lime} or gradient information~\cite{sundararajan2017ig, selvaraju2016gradcam}. Influential examples or influence function methods \cite{koh2017if, Garima2020estimating} instead quantify the effect of specific training examples on a given output. Concept extraction methods \cite{kim2017tcav, kazhdan2020cme} seek to measure a DNN's reliance on a set of understandable concepts. These methods are naturally interpretable and extendable to many DNN architectures~\cite{kazhdan2020meme, magister2021gcexplainer}.

Recent work has called into question the effectiveness of post-hoc explainers in both adversarial \cite{ kindermans2017reliability, dimanov2020trustme} and non-adversarial \cite{ghorbani2017fragile, basu2020iffragile, locatello2018assumpt, zhou2021attribute} settings. Given these deficiencies and their widespread usage, systematic methods of comparing and evaluating post-hoc explanations have became increasingly needed.

\paragraph{Evaluating Explainers}

There is no generally agreed method for comparing and evaluating post-hoc explainers. The majority of previous work has focused on feature attribution methods, proposing metrics to measure desirable qualities about the attribution method \cite{alvarez2018robustness, dai2022fairness, zhou2021evaluating}. The metrics often rely on semi-synthetic datasets containing `ground truth' explanations that correspond to the presence of known spurious signals \cite{agarwal2022openxai, liu2021benchmarks}. Metrics for other explainers remain limited with few exceptions \cite{zarlenga2022on}, and human trials are often still the only viable approach.

The closest work to ours is \citet{adebayo2022posthoc} which formulates a paradigm for evaluating DNN explainers for the purpose of identifying spurious correlations. Similar to our work, they focus on analysing spurious correlations in settings where the spurious signal is not known ahead of time via comparing explainers from spurious and non-spurious models. However, their framework does not allow for the direct comparison between different types of explainers as their proposed quantities have different units for different types of explainers. To the best of our knowledge, this work is the only presentation of a method for evaluating a post-hoc explainer's ability to detect spurious correlations that is comparable across all types of explainers while remaining focused on the context where the spurious signal is unknown.

\section{Explainer Divergence Score}
\label{sec.methodology}

We motivate our approach by considering the setting where a user seeks to determine whether a given model depends on a spurious signal using a post-hoc explainer. They inspect an explanation generated from a model prediction and use it to predict whether the model is spurious or not. Similarly to \citet{adebayo2022posthoc}, we expect a high-quality explainer to generate very different explanations from spurious models compared to non-spurious models.

This can be framed as a binary classification problem, where a classifier outputs a binary label corresponding to a prediction of a model's dependence on a spurious signal based upon an explanation as input. The classifier under this formulation is a machine learning model that takes the place of the user, and is trained to distinguish between explanations generated by spurious models and explanations generated by non-spurious models. A visual summary of our approach can be found in Figure \ref{fig:EDS_visual}.

%, of which it is known we can manipulate to generate arbitrary explanations~\cite{pmlr-v119-anders20a}

Critically, the classifier is trained to distinguish between explanations generated by \textit{all} spurious and non-spurious models generated by a specified training strategy, instead of any individual pair. This allows the classifier to generalize to unseen models much like a human user would be expected to. We detail how we accomplish this in Section \ref{sec.edsexpsetup}.

EDS is defined as the performance of this binary classifier in predicting model spuriousness on explanations generated using unseen models - and can be interpreted as a measure of explainer quality.

We can view our trained binary classifier's loss as an estimate of the distance between the distribution of explanations from spurious and non-spurious models respectively. Assume we have a trained binary classifier $f_{\hat{\theta}}$ parameterized by $\theta \in \Theta$. We train this classifier by minimizing the loss $\ell$ consisting of the cross-entropy $H$ between the distribution represented by the output of the model and $Y | \mathrm{x}$, that is, the distribution $Y$ conditioned on the random variable $\mathrm{x}$ where $Y$ is the Bernoulli distribution of binary labels of a given explanation in our training set (whether the model that generated it is spurious or non-spurious) and $\mathrm{x}$ is a random variable distributed according to the mixture distribution $X$ of equally weighted explanations from both the spurious and non-spurious models. We then have:

\begin{align} \label{eq.trained_sim}
\ell(f_{\hat{\theta}}) =& \min_{\theta \in \Theta} \mathbb{E}_{\mathrm{x} \sim X} \left[ H\left(Y | \mathrm{x}, f_\theta (\mathrm{x})\right)\right] \\
=& 1 - D_{\mathrm{JS}} (X | \mathrm{y} = 0, X | \mathrm{y} = 1) \\
& \quad \quad + \min_{\theta \in \Theta} \mathbb{E}_{\mathrm{x} \sim X} \left[D_{\mathrm{KL}}\left(Y |\mathrm{x} \Vert f_\theta (\mathrm{x})\right)\right] 
\end{align}

Where $D_{\mathrm{JS}}$ represents the Jensen-Shannon Divergence and $D_{\mathrm{KL}}$ represents the Kullback–Leibler Divergence, and all quantities are measured in bits of entropy. The full derivation of this expression is present in the Supplementary Material~\ref{sec.supplement}. Ideally, for a well trained classifier $f_{\hat{\theta}}$ of sufficient expressiveness, we would expect the distribution represented by the output of our classifier $f_{\hat{\theta}} (\mathrm{x})$ to approximate the true distribution of $Y | \mathrm{x}$, meaning the Kullback-Leibler Divergence between them is close to $0$: 
\begin{align} \label{eq.kl}
 \mathbb{E}_{\mathrm{x} \sim X} \left[D_{\mathrm{KL}}\left(Y | \mathrm{x} \Vert f_{\hat{\theta}} (\mathrm{x})\right)\right]  \simeq 0
\end{align}
And thus:
\begin{align}
\ell(f_{\hat{\theta}}) \simeq 1 - D_{\mathrm{JS}} (X | \mathrm{y} = 0, X | \mathrm{y} = 1)
\end{align}

In which case the loss of our trained model can be seen as approximating the Jensen-Shannon Divergence between the distribution of explanations generated by spurious models and the distribution of explanations generated by non-spurious models. Moreover, as all quantities share the same unit (information), they are directly comparable across explainers.

In practice, sufficient classifier accuracy for Equation \ref{eq.kl} to hold appears to be uncommon, leading to an average loss that is unbounded above and difficult to estimate. Hence we define our EDS as the classification accuracy of the binary classifier instead. This has the added advantage of providing an interpretable baseline for our metrics - if the classifier can not do better than random guessing (EDS of 0.5), then the classifier has failed to capture any information in the explanations useful for determining model spuriousness and thus there's a very low likelihood the explainer captures any information about the spurious signal.  

%We observe that our approach can be viewed as an alternative to \citet{adebayo2022posthoc}, which constructs a number of metrics based on premise of similarity functions between explanations. Given some similarity function $S_d$ and explainer $e_{f_{\hat{\theta}}}$, multiple metrics are proposed of the form $S_d (e_{f_{\hat{\theta}}}(x_0), e_{f_{\hat{\theta}}}(x_1))$ where $x_0$ and $x_1$ are carefully chosen subsets of the dataset.

%Our method can be viewed as using Equation \ref{eq.trained_sim} as a similarity function with $X | y = 0 \sim e_{f_{\hat{\theta}}}(x_0)$ and $X | y = 1 \sim e_{f_{\hat{\theta}}}(x_1)$ - effectively operating as a trained similarity function.

\section{Experiments}

Using a similar setup to \citet{adebayo2022posthoc, yang2022rare}, we investigated three different types of spurious artifacts:
\begin{itemize}
    \item \textbf{Square} - a small square in the top left corner of the image
    \item \textbf{Stripe} - a vertical stripe 9 pixels from the left of the image
    \item \textbf{Noise} - uniform Gaussian noise applied to every pixel value of the image
\end{itemize}

Examples of each spurious artifact on both the dSprites and 3dshapes datasets are present in the Supplementary Material~\ref{sec.supplement}.

We experiment to determine the effect of the intensity of each spurious artifact on a model's spurious behaviour, and then trained models to maximize this spuriousness. The details of this experiment and overall model training procedure can be found in the Supplementary Material~\ref{sec.supplement}.

\subsection{EDS Experimental Setup}

%We evaluate our Binary Classifier Metric on both datasets as follows. We split each dataset into 80\% training and 20\% validation partitions. Then, using different weight initializations, we train 100 spurious and non-spurious models on the training subsets following the procedure outlined in Section \ref{sec.training} with specific details in Appendix \ref{sec.procedure}. Using a large number of models is critical as we want to draw conclusions about the entire distributions of spurious and non-spurious model weights rather then any single pair of trained models.

\label{sec.edsexpsetup}

For all datasets and explainers, we evaluate the Explainer Divergence Score (EDS) as follows. We split the dataset into three partitions - 80\% partition used for model training, 14\% partition used for binary classifier training, and 6\% partition used for validation.

Recall in Section \ref{sec.methodology} we defined EDS using a binary classifier trained to distinguish between explanations generated across all spurious and non-spurious models. Training a new model for every explanation is far too computationally intensive. To rectify this for each spurious artifact we train 100 spurious and 100 non-spurious models on our model training dataset partition, using different weight initialization, and use this sample as an estimate of the complete distribution of trained spurious and non-spurious models respectively. We train models and ensure they are spurious or non-spurious respectively using the procedure detailed in the Supplementary Material~\ref{sec.supplement}.

We reserve 30 spurious and non-spurious models each for validation and use the remaining 70 of each set to generate training data for our binary classifier. Images from the respective dataset partition are combined with a randomly selected model to generate an explanation as well as a binary class label corresponding to whether the model came from a spurious or non-spurious set. A classifier is then trained on this data to use the explanations to predict this class label.

Finally, our remaining 30 spurious and 30 non-spurious models are combined with the validation dataset partition to generate explanations in the same fashion as in training. The label prediction accuracy of the binary classifier on this set is then our estimate of the Explainer Divergence Score of the given explainer for this spurious signal. Further experimental setup details are noted in the Supplementary Material~\ref{sec.supplement}.

%We then split our validation partition again into 70\% and 30\% subpartitions, and randomly select 70 spurious and non-spurious models respectively to train the binary explanation classifier. We reserve 30 models of each set to evaluate classifier accuracy. We do this as our explanations are functions of both a given model and a given image, so we evaluate our binary classifier on both unseen models and unseen images.

%We now generate classifier training and validation explanations from the models and images in their respective subpartitions. We assign binary class labels to each explanation indicating whether it was generated using a spurious or a non-spurious model. Finally, we train our binary classifier to predict these class labels from the training explanations, and evaluate its accuracy on the validation explanations subset. %This process is illustrated in Figure [PLACEHOLDER].

\subsection{Subclass Definitions}
\label{sec.subclasses}

For all EDS results we display accuracy not just over the entire dataset (noted as `Overall' in figures), but also subdivided by the task label and the presence of the spurious artifact in the image. There are four subclasses in total:
\begin{itemize}
    \item Images from the Spurious Class without the Spurious Artifact (abbreviated as `S/NA' in figures)
    \item Images from Non-Spurious Classes without the Spurious Artifact (abbreviated as `NS/NA' in figures)
    \item Images from the Spurious Class with the Spurious Artifact (abbreviated as `S/A' in figures)
    \item Images from Non-Spurious Classes with the Spurious Artifact (abbreviated as `NS/A' in figures)
\end{itemize}
For example, say we had a class consisting of images of `circles' and another of images of `squares', and we trained a classification model between the two where we injected spurious Gaussian noise into the `circles' class. `S/NA' would correspond to images of circles without Gaussian noise, `NS/NA' would correspond to images of squares without Gaussian noise, `S/A' would correspond to images of circles with Gaussian noise, and `NS/A' would correspond to images of squares with Gaussian noise.

This subdivision allows us to interpret the type of images the explainer can effectively use to determine model spuriousness. This is analogous to what is done in \citet{adebayo2022posthoc} via the `Cause-for-Concern Metric' (CCM) and `False Alarm Metric' (FAM) that measure results by whether the spurious artifact is present in the image, but we present results in even finer detail with added class information.

\subsection{Synthetic Explainer Comparison}

 We compare our EDS method to the approach in \citet{adebayo2022posthoc}, starting with a simple example. We consider a toy classification task with two simple classes (the dSprites classes of a `heart' and `oval') with the `stripe' spurious artifact injected. Instead of using a specific spurious detection method, we instead construct synthetic explainers that represent the expected behaviour of each method under ideal circumstances. We construct these `ideal' explainers as follows:

\begin{itemize}
    \item \textbf{Heatmaps} - for the spurious model the explainer places all emphasis on the stripe for all images where it is present and the area where the stripe \textit{would be} for images from the spurious class without the stripe. The explainer puts all emphasis on the shape for all cases with the non-spurious model and for the spurious model on non-spurious classes without the stripe.
    \item \textbf{Influential Examples} - for the spurious model the explainer selects influential examples of the spurious class with the stripe for all images unless it is an image from a non-spurious class without the stripe. For the non-spurious model the explainer always selects examples of the correct class with the correct presence of the spurious artifact.
    \item \textbf{Concept Extraction} - For concept extraction we specify two binary concepts, one of the class label and one of the presence of the spurious artifact. We assume the spurious model can detect both perfectly, and thus extracts both accurately. On the other hand, the non-spurious model is invariant to spurious artifact in all circumstances, and thus always extracts that it is not present.
\end{itemize}

In addition to these ideal explainers, we also create `noisy' variants where we inject noise across every explanation as well as a purely random variant where the corresponding explanations consist purely of noise. For heatmaps we inject uniform Gaussian noise to the heatmap, for influential examples we specify a chance (100\% in the noise variant) of randomly selecting a training image, and for concept extraction we specify a chance (100\% in the noise variant) of predicting a random concept label. 

We evaluate both our EDS and the KSSD, CCM and FAM metrics \cite{adebayo2022posthoc} on these examples. For these metrics we specify similarity functions as follows: for heatmaps we use the SSIM similarity function as specified in \cite{adebayo2022posthoc}, for influential examples we use the Bhattacharyya coefficient \cite{kailath1967bhattacharyya} between the distributions of the class labels and the presence of a spurious artifact in the influential examples, and for concept extraction we use the negative of the L2 distance between concept labels as a `similarity' function. The results are shown in Table \ref{tab:synthheat}.

%\begin{table}[]
%\begin{tabular}{|l|l|llll|ll|}
%\hline
%Explainers & Overall & S/NA & NS/NA & S/A & NS/A & CCM & FAM \\ %\hline
%Heatmap & 0.875 &   1.0  & 0.5  &  1.0 &  1.0 & 0.993 & 0.986 \\ %\cline{7-8} 
%Influence & 0.875&   1.0 &   0.5  &  1.0 &  1.0 & 0.5 & 0.0 \\ %\cline{7-8} 
%Concept & 0.75 &   0.5 &   0.5 &  1.0 &  1.0 & 0.0 & -0.500 \\  %\hline
%\end{tabular}
%\caption{Comparison between Perfect Synthetic Explainers}
%\label{tab:synthexp}
%\end{table}

\begin{table*}[]
\begin{tabular}{|l|l|llll|lll|}
\hline
Explainers &  \multicolumn{5}{c|}{Explainer Divergence Scores} & \multicolumn{3}{c|}{\citet{adebayo2022posthoc} Metrics}  \\ \hline
 & Overall & S/NA & NS/NA & S/A & NS/A & KSSD & CCM & FAM  \\ \hline
Heatmaps & & &  & & & & &  \\ \hline
Ideal & 0.823 & 0.943 & 0.492 & 0.959 & 0.896 & 1.000 & 0.991 & 0.990 \\
Noisy &  0.702 & 0.771 & 0.459 & 0.805 & 0.771 & 0.970 & 0.993 & 0.993 \\
Random & 0.512 & 0.518 & 0.510 & 0.537 & 0.484 & 0.965 & 0.996 & 0.996 \\ \hline
Influence &  & & & & &&  &  \\ \hline
Ideal & 0.824 & 0.955 & 0.496 & 0.949 & 0.896 & 1.000 & 0.500 & 0.000  \\
Noisy & 0.668 & 0.734 & 0.490 & 0.736 & 0.713 & 0.587 & 0.712 & 0.656 \\
Random & 0.515 & 0.510 & 0.516 & 0.520 & 0.514 & 0.000 & 0.915 & 0.927 \\ \hline
Concept & & & & & &  &  &\\ \hline
Ideal & 0.750 & 0.500 & 0.500 & 1.000 & 1.000 & 0.000 & 0.000 & -0.500 \\
Noisy & 0.617 & 0.488 & 0.535 & 0.723 & 0.721 & -0.284 & -0.412 & -0.514 \\
Random & 0.491 & 0.490 & 0.475 & 0.527 & 0.473 & -0.494 & -0.481 & -0.509 \\ \hline

\end{tabular}
\caption{Results of evaluating EDS on the specific synthetic explainers averaged over 5 runs. We observe clear outperformance of heatmaps and influential examples over concept extraction, as well as the complete failure of the `random' explainer in the EDS results. However, these are not visible in the KSSD, CCM and FAM metrics~\cite{adebayo2022posthoc} results. Standard deviation estimates are provided in the Supplementary Material~\ref{sec.supplement}, with all results having estimated 95\% confidence intervals within $\pm 0.03$.}
\label{tab:synthheat}
\end{table*}

Synthetic `ideal' explainers are useful as we can specify in advance exactly how our explainers should perform between the spurious and non-spurious models in each subclass. Cases where the explanations generated from spurious and non-spurious models are drawn from the same distribution should result in the worst possible metrics. Conversely, cases where the explanations are always radically different should result in close to perfect metrics. 

This is exactly what we observe with EDS. Our approach finds the ideal heatmap and influential examples almost perfectly identify model spuriousness - failing only on explanations generated from images from a non-spurious class without the spurious artifact. The ideal concept extraction explainer additionally falls short on images from the spurious class with the spurious artifact, indicating that this specification is a worse explainer for detecting spurious correlations then the competing methods.

We observe that the KSSD, CCM and FAM metrics from \citet{adebayo2022posthoc} fall short in this type of analysis: different types of explainers use different similarity functions with different units that are not comparable directly. This is a major innovation of our method over the existing state of the art.

Our method comes to our expected conclusion that the ideal explainers capture more information about model spuriousness than the noisy explainers, while the random explainers completely fail to capture any information about model spuriousness. This declining performance can also be seen in the KSSD, CCM and FAM metrics - but the utter failure of the random explainers is not visible with these metrics. With EDS, if the trained classifier fails to achieve at least 50\% accuracy, we can interpret the explainer as having no information about the model's spuriousness. This is not possible using the KSSD, CCM and FAM metrics without explicitly running a baseline for every type of explainer evaluated.

\subsection{Real Explainer Comparison}

To test EDS on real explainer methods, we conduct experiments on reduced versions of both the dSprites \cite{dsprites17} and the 3dshapes \cite{3dshapes18} datasets. We train models to perform a shape classification task and arbitrarily select one class to be the spurious class for each experiment.

We chose some commonly used methods as representatives for each explainer type of interest. We use Integrated Gradients \cite{sundararajan2017ig} as our chosen feature attribution method. For influential examples we use the TraceInCP method \cite{Garima2020estimating}, and for concept extraction we use Concept Model Extraction (CME) \cite{kazhdan2020now}. Examples of each explanation on images from the dSprites dataset are present in the Supplementary Material~\ref{sec.supplement}.

More detailed information about the configuration setup for each experiment is present in the Supplementary Material~\ref{sec.supplement}.

We display results for dSprites in Table \ref{tab:dsprites_bcm} and results for 3dshapes in Table \ref{tab:3dshapes_bcm}. For comparison, we also evaluate the KSSD, CCM, and FAM metrics formulated in \citet{adebayo2022posthoc} on both dSprites and 3dshapes.

\begin{table*}[]
\begin{tabular}{|l|l|llll|lll|}
\hline
Explainers &  \multicolumn{5}{c|}{Explainer Divergence Scores} & \multicolumn{3}{c|}{\citet{adebayo2022posthoc} Metrics}  \\ \hline
 & Overall & S/NA & NS/NA & S/A & NS/A & KSSD & CCM & FAM  \\ \hline
 Square & & &  & & & & & \\ \hline
Heatmap & 0.799 & 0.837 & 0.590 & 0.902 & 0.916 & 0.851 & 0.877 & 0.837     \\
Influence & 0.887 & 0.937 & 0.860 & 0.891 & 0.887 & 0.562 & 0.991 & 0.989\\
Concept & 0.715 & 0.645 & 0.578 & 0.827 & 0.831 & -0.062 & -0.074 & -0.076\\ \hline
Stripe  & & &  & & & & & \\ \hline
Heatmap & 0.831 & 0.901 & 0.689 & 0.958 & 0.870 & 0.877 & 0.880 & 0.878 \\
Influence & 0.881 & 0.892 & 0.829 & 0.909  & 0.913 & 0.561 &  0.991 & 0.980\\
Concept & 0.707 & 0.618 & 0.596 & 0.788 & 0.815 & -0.061 & -0.074 & -0.077\\ \hline
Noise & & &  & & & & &  \\ \hline
Heatmap & 0.717 & 0.857 & 0.610 & 0.872 & 0.682 & 0.728 & 0.877 & 0.804 \\
Influence & 0.795 & 0.884 & 0.707 & 0.890 & 0.796 & 0.566 & 0.970 & 0.966\\
Concept & 0.744 & 0.650 & 0.652 & 0.863 & 0.808 & -0.062 & -0.078 & -0.076\\ \hline

\end{tabular}
\caption{Results of evaluating EDS on the dSprites dataset  averaged over 5 runs. We observe the outperformance of influential examples over heatmaps and concept extraction visible in the EDS results but not in the comparative metrics. Standard deviation estimates are provided in the Supplementary Material~\ref{sec.supplement}., with all results having estimated 95\% confidence intervals within $\pm 0.04$.}
\label{tab:dsprites_bcm}
\end{table*}

\begin{table*}[]
\begin{tabular}{|l|l|llll|lll|}
\hline
Explainers &  \multicolumn{5}{c|}{Explainer Divergence Scores} & \multicolumn{3}{c|}{\citet{adebayo2022posthoc} Metrics}  \\ \hline
 & Overall & S/NA & NS/NA & S/A & NS/A & KSSD & CCM & FAM  \\ \hline
 Square & & &  & & & & &  \\ \hline
Heatmap & 1.00 & 1.00 & 1.00 & 1.00 & 1.00 & 0.680 & 0.828 & 0.826   \\
Influence & 0.675 & 0.817 & 0.615 & 0.696  & 0.682 & 0.562 & 0.991 & 0.989\\
Concept & 0.595 & 0.532 & 0.562 & 0.651  & 0.628 & -0.156 & -0.080 & -0.072\\ \hline
Stripe & & &  & & & & &  \\ \hline
Heatmap & 0.996 & 0.993 & 0.997 & 0.998  & 0.994 & 0.644 & 0.810 & 0.807  \\
Influence & 0.867 & 0.922 & 0.844 & 0.903 & 0.861 & 0.561 & 0.991 & 0.980\\
Concept & 0.720 &  0.653 & 0.636 & 0.783  & 0.795 & -0.152 & -0.075 & -0.070\\ \hline
Noise & & &  & & & & &  \\ \hline
Heatmap & 0.987 & 1.00 & 0.984 & 0.994 & 0.984 & 0.673 & 0.846 & 0.847 \\
Influence & 0.703 & 0.908 & 0.645 & 0.887 & 0.627 & 0.566 & 0.970 & 0.966\\
Concept & 0.569 & 0.600 & 0.566 & 0.587 & 0.555 & -0.150 & -0.074 & -0.074\\ \hline

\end{tabular}
\caption{Results of evaluating EDS on the 3dshapes dataset  averaged over 5 runs. We observe the extreme outperformance of heatmaps over the remaining explainers visible in the EDS results but not in the comparative metrics. Standard deviation estimates are provided in the Supplementary Material~\ref{sec.supplement}, with all results having estimated 95\% confidence intervals within $\pm 0.04$.}
\label{tab:3dshapes_bcm}
\end{table*}

We observe Explainer Divergence Scores significantly above the 0.5 theoretical baseline for all explainers and spurious artifacts in both datasets. This indicates all of our explainers are successful in capturing information about the model's spuriousness in both tasks. The key advantage of EDS over previous work is that we can now directly compare the performance of explainers for detecting model spuriousness for the specified task and spurious artifact. We interpret our results with this aim in mind.

We find the strongest performance for heatmaps and influential examples. EDS was highest for images in the spurious class without the spurious artifact, lowest for images in non-spurious classes without the spurious artifact, and somewhat high for images with the spurious artifact regardless of class. These findings appear consistent across all three of our chosen spurious artifacts, and in both datasets. We notice a sharp drop in performance for our Gaussian noise spurious artifact compared to the more localized spurious artifacts. 

Concept extractions consistently perform worse than the other two explainers, operating well only on images with the explicit presence of the spurious artifact. This follows our expectations - we would expect concept extraction to more effectively identify the presence of the spurious signal concept from the activations of spurious models compared to activations of non-spurious models that have learned to become invariant to them. Moreover the dimensionality of our the concept predictions is much lower than explanations for the other two explainers, limiting their expressiveness. Interestingly while performance on images without the spurious artifact is poor, it is still above our 0.5 theoretical baseline despite there being no obvious reason for concept predictions to shift between spurious and non-spurious models. This is further discussed in Section \ref{sec.shifts}.

We notice significant differences in explainer performance between the dSprites and 3dshapes datasets. While in dSprites we find slightly higher performance for influential examples over heatmaps, in 3dshapes we find significant strong performance for heatmaps across all experiments. In 3dshapes often our EDS binary classifier identifies the spuriousness of a given model from a heatmap with 100\% accuracy. This is in sharp contrast to the other two explainers that perform worse with 3dshapes, performing only comparably using the `stripe' spurious artifact. Despite this diminished performance, both influential examples and concept extraction still perform above our 0.5 theoretical baseline for EDS. %The significant variation in our results between the 'stripe' and other spurious artifacts may suggest the artifact is affecting the ability of the model to identify class information beyond just the effect of the spuriousness, possible by occluding the left edges of the central shape in certain orientations.

\subsection{Discussion}
\label{sec.shifts}

These results favour heatmaps and influential examples, which are very effective at detecting model spuriousness in both experiments with real explainer methods. Conversely concept extraction consistently performed the worst, and is only useful on images for which the spurious artifact is present. As expected, performance is sensitive to the dataset and specified task. 

We conduct further experiments to confirm Explainer Divergence Scores are robust to our choice of optimization procedure and model architecture. These are expanded upon in the Supplementary Material~\ref{sec.supplement}.

In both datasets, we observe EDS performances significantly above the 0.5 theoretical baseline for all explainers, spurious artifacts, and subclasses. Notably this is seen even with images from unrelated, non-spurious classes without the presence of the spurious artifact. 

This has interesting implications about the utility of post-hoc explainers in detecting model spuriousness. For example, heatmaps generated from 3dshapes images in non-spurious classes  without the spurious artifact do not show any obvious signal that a human could use to identify their respective model has some sort of spurious dependency. Yet a trained classifier with sufficient prior knowledge can diagnose whether the model depends upon a spurious signal with extremely high certainty. Information present in our explanations indicating spuriousness may not always be perceptible by a human observer, and identifying ways to extract or isolate this information may prove useful in designing more effective explainers. 

This is particularly evident in the case of concept extraction where there is no clear hypothesis for why spurious and non-spurious models would have differing information about the underlying concepts in images from the non-spurious class without the spurious artifact. This suggests that the presence of a spurious correlation can affect a model's ability to extract features in entirely unrelated image classes.

\section{Conclusion}

We present Explainer Divergence Scores - a novel method for evaluating post-hoc explainers for the purposes of detecting unknown spurious correlations.

Across three experiments we show EDS's superior capabilities over state of the art post-hoc explainer evaluation methods. EDS provides an interpretable estimate of the amount of information an explainer can capture about a DNN's dependence on an unknown spurious signal. Moreover EDS allows direct comparisons between different types of explainers, unlike previous methods, letting us quantitatively identify and evaluate the best explainer for a given dataset and spurious signal.

In contrast to previous work~\cite{adebayo2022posthoc}, our results reveal that commonly used post-hoc explainers contain substantial amount of information about a model's dependence on unknown spurious signals. This information is often unidentifiable by human observers, and yet can be used by a well-trained classifier to detect dependencies on images seemingly unrelated to the spurious signal. Our findings suggest that future research into post-hoc explanations should focus on identifying and utilizing this unseen information. 

\section{Supplementary Material}
\label{sec.supplement}

Additional information about our work, including a more detailed mathematical justification, ancillary experiments, and standard error estimates for all our results are detailed in the Appendix available at \href{https://bit.ly/3emRR9j}{this link}.

%%
%% Define the bibliography file to be used
\bibliography{references}

\end{document}

%% file: Images/abstract_visual.tex
\begin{figure*}
  \begin{center}
        \centering
      \includegraphics[scale=0.55] 
      {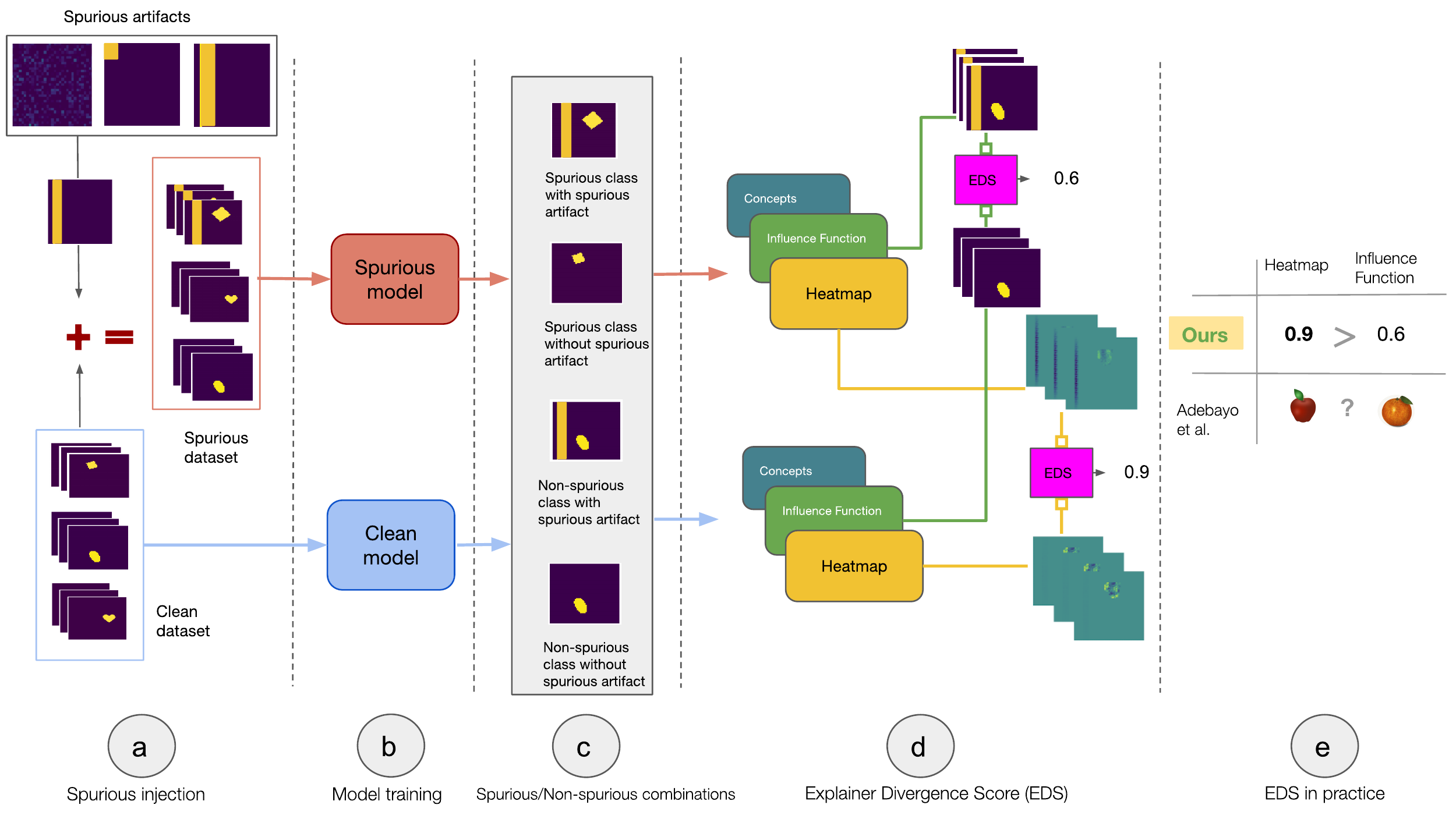}
  \end{center}
  \caption{Explainer Divergence Score (EDS). \protect\encircle{a} With one engineered spurious dataset and one clean dataset, \protect\encircle{b} we train two separate classification models. \protect\encircle{c} These models evaluate different combinations of spurious and non-spurious examples. \protect\encircle{d} EDS can assess a post-hoc explainer's ability to detect spurious correlations. \protect\encircle{e} In comparison to previous work, our approach allows us to compare the performance of different types of post-hoc explainers directly.}
  \label{fig:EDS_visual}
\end{figure*}